%% file: main.tex
\documentclass[letterpaper]{article} 
\usepackage[preprint]{aaai2027}  
\usepackage[hyphens]{url}  
\usepackage{graphicx}  
\usepackage{natbib}  
\usepackage{caption} 
\usepackage{hyperref}
\usepackage{algorithm}
\usepackage{algorithmic}
\usepackage{array}
\usepackage{multirow}
\usepackage{newfloat}
\usepackage{listings}
\DeclareCaptionStyle{ruled}{labelfont=normalfont,labelsep=colon,strut=off} 
\floatstyle{ruled}
\newfloat{listing}{tb}{lst}{}
\floatname{listing}{Listing}

\usepackage{booktabs}
\usepackage{amsmath}

\title{To See a World in a Living Context:\\Unified Indoor-Outdoor Urban World Generation}

\author{
    Xiaobin Huang\textsuperscript{1}\equalcontrib,
    Zilong Huang\textsuperscript{1}\equalcontrib,
    Yang Luo\textsuperscript{1},
    Hongchao Fan\textsuperscript{2},
    Yiping Chen\textsuperscript{1}\corresponding,
    Ting Han\textsuperscript{1}\corresponding
}

\affiliations{
    \textsuperscript{1}Sun Yat-sen University \\
    \textsuperscript{2}Norwegian University of Science and Technology 
}

\begin{document}

\twocolumn
[{
\renewcommand\twocolumn[1][]{#1}
\maketitle
\begin{center}
\vspace{-1cm}
\captionsetup{type=figure}
\includegraphics[width=\textwidth]{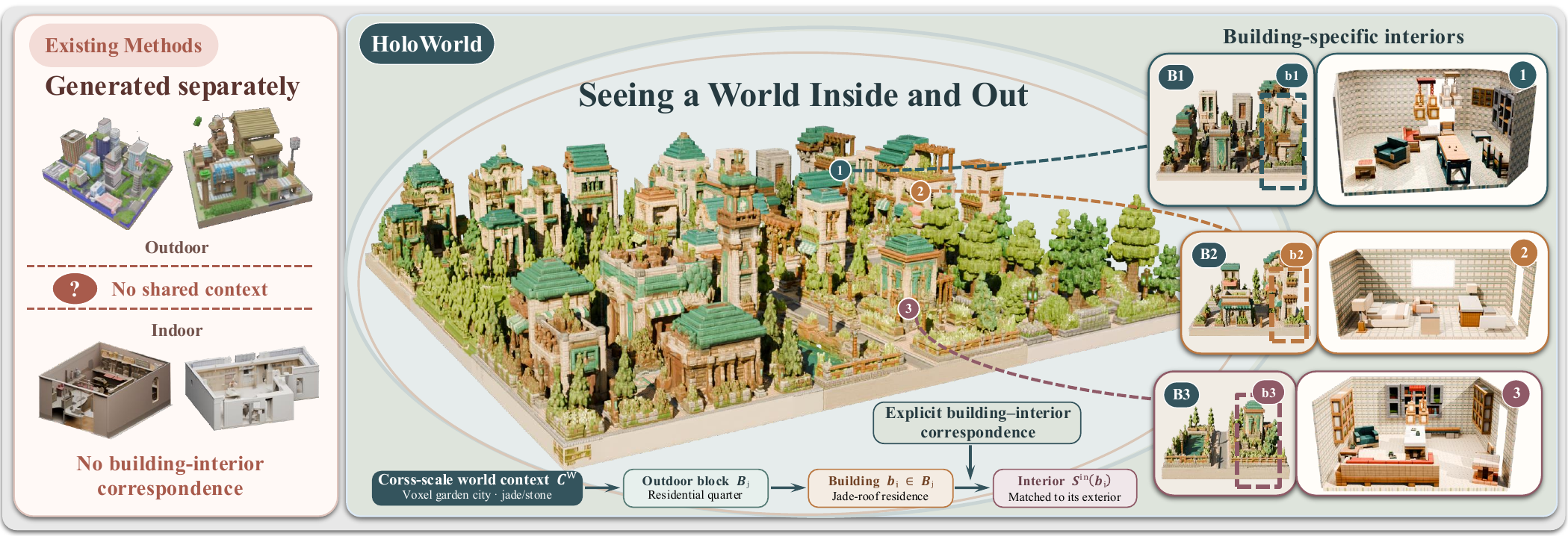}
\caption{Existing methods generate indoor and outdoor scenes separately. HoloWorld instead propagates a cross-scale world context from the urban world through its blocks to individual buildings, establishing explicit building--interior correspondence and preserving semantic, visual, and spatial coherence.}
\label{fig:teaser}
\end{center}
}]

\input{sec/abstract}

\input{sec/introduction}

\input{sec/related_work}

\input{sec/method}

\input{sec/experiment}

\input{sec/conclusion}

\bibliography{aaai2027}

\end{document}

%% file: sec/abstract.tex
\begin{abstract}
Text-driven 3D generation has advanced rapidly in creating large-scale outdoor environments and detailed indoor scenes, but these domains are usually synthesized independently, lacking the correspondence required for a coherent urban world. We present HoloWorld, a unified indoor-outdoor urban world generation framework built on a continuously updated cross-scale world context. Initializing from a user description, HoloWorld progressively represents and updates the diverse world information, from city-scale planning to individual buildings, allowing generated interiors to maintain explicit correspondence with their associated exterior buildings. Conditioned on the evolving context and previously generated neighboring blocks, HoloWorld autoregressively generates urban exteriors with consistent spatial organization and visual identity across blocks. The generated exterior representations are further grounded in 3D building instances and footprints, enabling building-specific indoor generation with geometry-constrained layouts and inherited appearance characteristics. To our knowledge, HoloWorld is the first framework to unify indoor and outdoor generation within a coherent 3D urban world. Extensive experiments demonstrate that HoloWorld achieves superior urban exterior generation performance, improving the average AQS score over the SOTA by 7.68\% and obtaining the highest average RDR score, while maintaining strong building-level indoor-outdoor correspondence and cross-block continuity within a unified 3D urban world.
Our project page: \href{blue}{https://huangxb326.github.io/HoloWorld/}
\end{abstract}

%% file: sec/introduction.tex
\section{Introduction}
\label{sec:introduction}

\textit{``To see a World in a Grain of Sand / And a Heaven in a Wild Flower''}~\cite{blake_auguries}. Blake's lines describe how a complete world can be perceived through a local fragment, highlighting a fundamental property of coherent environments: local observations should remain consistent with a larger global identity. This principle is increasingly important for immersive virtual worlds, embodied-agent simulation~\cite{procthor}, and interactive 3D applications, where users and agents continuously navigate across spatial scales. However, existing text-driven 3D generation methods~\cite{majutsucity,yocity,holodeck,mansion} often produce visually plausible scenes that are disconnected from one another: An urban exterior may appear realistic, but its interior often does not match the same world when generated separately. Therefore, a coherent virtual world requires preserving semantic, visual, and geometric consistency across scales.

Recent advances in urban-scale 3D generation have enabled the synthesis of large environments with diverse architectures, spatial layouts, and realistic appearances~\cite{urbangiraffe,infinicity,citydreamer,gaussiancity,syncity,yocity}. Meanwhile, indoor scene generation has achieved remarkable progress in language-guided room planning, asset synthesis, and hierarchical layout construction~\cite{atiss,holodeck,mansion,scenesmith}. Nevertheless, these two research directions remain fundamentally separated. Urban generation methods primarily focus on constructing exterior environments without modeling the internal spaces of individual buildings, whereas indoor generation methods typically generate isolated scenes without grounding them in an existing urban context. Even recent general agent-based 3D generation systems rarely establish an explicit correspondence between a generated building and its interior realization~\cite{scenecraft,worldcraft,scenethesis}.

We believe that a unified indoor-outdoor generation framework should construct a plausible and coherent urban world in which every interior is explicitly grounded in its exterior counterpart. Specifically, each generated interior should correspond to a building instance, inherit the building's functional semantics and visual identity, and respect its footprint. Such a formulation transforms indoor-outdoor generation from independent scene synthesis into a world-consistent generation problem, requiring cross-scale reasoning and information propagation throughout generation.

The key challenge is to preserve, propagate, and localize contextual information across spatial scales during generation. City-level and block-level generation establishes global semantic structures and visual styles, whereas indoor synthesis requires localized conditions associated with individual building instances and their geometries. Therefore, a unified framework needs a structured world representation and state that can represent the semantic, visual, and geometric information from the city scale to the building scale. 

To address this challenge, we introduce \textbf{HoloWorld}, a unified indoor-outdoor urban scene generation framework built upon a cross-scale world context (Figure~\ref{fig:teaser}). The world context serves as a shared representation that records and transfers validated semantic, visual, spatial, and content information throughout the generation. Initialized from the user description, it is progressively refined from the city level to blocks and individual buildings, allowing generated interiors to inherit relevant knowledge from their surrounding urban environments. To establish correspondence between outdoor and indoor spaces, HoloWorld employs a cross-scale context bridging mechanism that transfers semantic and visual information from textual intent to world realization, maintains continuity across neighboring urban regions, and grounds exterior building structures to corresponding interiors. Based on the resulting building-level context, HoloWorld generates interiors that are not only visually plausible but also functionally aligned and geometrically grounded with their corresponding exterior buildings. 

Extensive experiments on diverse urban scenarios demonstrate that HoloWorld achieves superior indoor-outdoor consistency and urban generation quality compared with existing approaches, improving the average AQS score over the SOTA baseline by 7.68\% and obtaining the highest average RDR score. To our knowledge, HoloWorld represents a first step toward moving beyond isolated indoor and outdoor synthesis by generating corresponding spaces as a unified and coherent 3D urban world. The main contributions are as follows:
\begin{itemize}
    \item We formulate the task of unified indoor-outdoor urban scene generation and introduce HoloWorld, which generates corresponding indoor and outdoor spaces as a coherent 3D urban world.
    \item We propose a cross-scale world context that enables coherent autoregressive generation across blocks and consistent correspondence between urban exteriors and building interiors.
    \item We develop a building-grounded indoor generation strategy that concentrates interior synthesis on exterior building instances, enabling unified indoor-outdoor world generation.
\end{itemize}

%% file: sec/related_work.tex
\section{Related Work}
\label{sec:related-work}

Urban-scale 3D generation must coordinate spatial organization, architectural diversity, and visual consistency across large regions. CityDreamer models unbounded cities through compositional representations of building instances and background elements~\cite{citydreamer}. Language-guided systems further combine layout generation, urban planning, and controllable asset assembly~\cite{citycraft,majutsucity}. SynCity expands a world tile by tile while conditioning each new region on previously generated surroundings~\cite{syncity}, whereas Yo'City uses an agentic framework to support personalized and spatially coherent city growth~\cite{yocity}. These methods substantially improve the structure, controllability, and continuity of urban exteriors, but remain centered on exterior world construction rather than carrying city-level semantics, style, and generated evidence forward to the interiors of specific buildings.

Indoor generation has progressed from learned furniture arrangement~\cite{atiss,sceneformer,diffuscene} to language-driven multi-room and building-scale synthesis~\cite{mansion,anyhome}. Holodeck, SAGE, and SceneSmith further use language or vision-language agents for content planning, iterative refinement, and simulator-ready scene construction~\cite{holodeck,sage,scenesmith}. General systems broaden text-driven generation across scene types through unified representations, procedural construction, code synthesis, or visual feedback~\cite{3dscenedreamer,text2nerf,worldgen,threedgpt,scenewiz3d,scenecraft,scenethesis,sceneassistant}. WorldCraft applies coordinated language agents and procedural tools to indoor and outdoor scene design~\cite{worldcraft}, but does not formulate the correspondence between an urban building and its interior. ShellMaker instead completes an exterior from a prescribed structural scaffold while preserving its footprint, walls, and openings~\cite{shellmaker}. Together, these works expand the scope of 3D generation, but supporting both domains does not establish a traceable building-to-interior correspondence or preserve semantic, visual, and spatial context across the building boundary.

%% file: sec/method.tex
\section{Method}
\label{sec:method}

\begin{figure*}
    \centering
    \includegraphics[width=\linewidth]{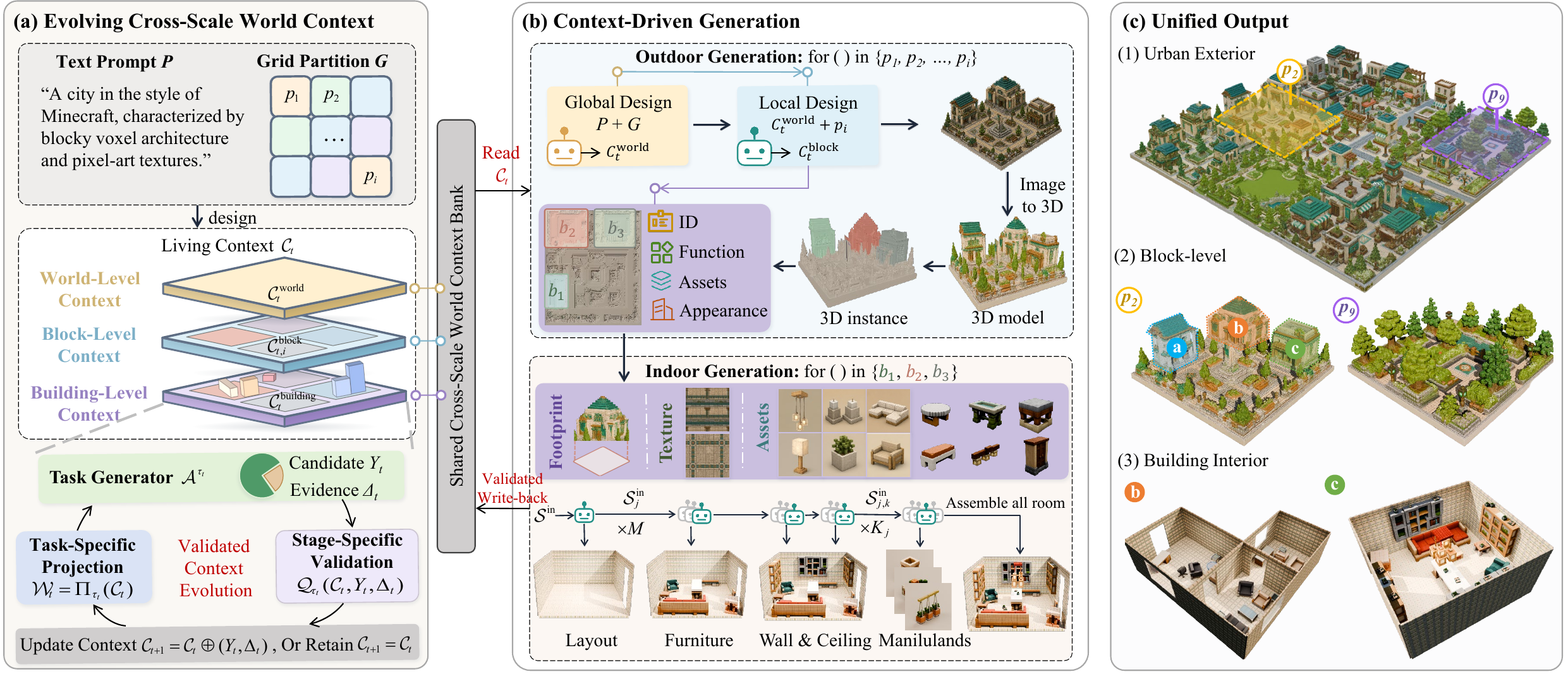}
    \caption{Overview of HoloWorld. (a) An evolving cross-scale world context organizes validated information at world, block, and building levels. (b) Context-driven outdoor generation and instance grounding localize this context to individual buildings, where it guides building-specific indoor synthesis under appearance, asset, and footprint constraints. (c) The resulting city, blocks, and interiors form explicitly corresponding parts of a unified 3D urban world.}
    \label{fig:method-overview}
    \vspace{-0.4cm}
\end{figure*}

\subsection{Problem Formulation}
Figure~\ref{fig:method-overview} summarizes the HoloWorld pipeline, from its evolving cross-scale world context and context-driven generation to the unified indoor-outdoor output. We formulate unified indoor-outdoor urban scene generation as a cross-scale 3D world generation task that jointly models urban exteriors and their corresponding building interiors. Given an arbitrary text prompt $P$ describing an urban intent, our objective is to generate a coherent 3D urban world in which spatially organized exteriors and their corresponding interiors maintain consistent semantic and geometric relationships.

We represent the urban exterior as an $R\times C$ grid of blocks $G=\{0,\ldots,R-1\}\times\{0,\ldots,C-1\}$, where each block denotes a spatial unit in the generated urban world. The collection of blocks defines the urban exterior $\mathcal{S}^{\mathrm{out}}$, within which generated building instances form a set $\mathcal{B}$ with stable identities. Let $\mathcal{B}_{\mathrm{tar}}\subseteq\mathcal{B}$ denote the subset of buildings selected for indoor generation, and let $\mathcal{S}^{\mathrm{in}}_b$ represent the interior associated with building instance $b$. The unified output is defined as:
\begin{equation}
\mathcal{O}
=
\left(
\mathcal{S}^{\mathrm{out}},
\left\{
(b,\mathcal{S}^{\mathrm{in}}_b)
\mid b\in\mathcal{B}_{\mathrm{tar}}
\right\}
\right).
\label{eq:unified-output}
\end{equation}

Therefore, HoloWorld aims to generate indoor and outdoor spaces as consistent representations of the 3D urban world. The key idea is to maintain a cross-scale world context $\mathcal{C}$ that bridges textual semantics, visual realization, geometric grounding, and indoor-outdoor generation across spatial scales. Starting from the user prompt, the context evolves from global urban planning to individual building generation, progressively refining the representation of the generated world.

Unlike independent scene synthesis, our formulation requires consistency at both urban and building scales. At the urban scale, the context guides block generation and preserves spatial continuity between neighboring regions. At the building scale, the context associates each generated interior $\mathcal{S}^{\mathrm{in}}_b$ explicitly with the corresponding exterior building instance $b$, inheriting the building's semantics and appearance while remaining constrained by its geometric footprint. Through this evolving context, HoloWorld jointly generates coherent urban exteriors and building-grounded interiors within a unified 3D world.

\subsection{Cross-Scale World Context Representation}

To maintain a consistent identity of the generated world across spatial scales, we represent the generation process through a hierarchical cross-scale world context $\mathcal{C}$. Unlike a simple memory that stores previous outputs, $\mathcal{C}$ serves as a structured representation that bridges semantic descriptions, visual appearances, spatial layouts, and building geometries throughout generation.

At the generation stage $t$, the world context is decomposed into three hierarchical levels:
\begin{equation}
\mathcal{C}_t
=
\left(
\mathcal{C}^{\mathrm{world}}_t,
\left\{\mathcal{C}^{\mathrm{block}}_{t,i}\right\}_{i\in G},
\left\{\mathcal{C}^{b}_t\right\}_{b\in\mathcal{B}_t}
\right).
\label{eq:world-context}
\end{equation}
where $\mathcal{B}_t$ denotes the set of building instances grounded at stage $t$.

The world-level context $\mathcal{C}^{\mathrm{world}}_t$ encodes global information that defines the identity of the generated city, including urban semantics, functional organization, and shared visual styles. The block-level context $\mathcal{C}^{\mathrm{block}}_{t,i}$ specializes this global information for individual spatial regions and maintains the local spatial relationships required for coherent block generation. The building-level context $\mathcal{C}^{b}_t$ further localizes the inherited information to a specific building instance by integrating its identity, appearance, and geometric properties, providing the conditions required for corresponding indoor synthesis.

These three levels form a hierarchical representation rather than independent states. Information is inherited from higher levels to lower levels, while newly generated and validated evidence is propagated back to enrich the corresponding context. This bidirectional interaction enables the generated city to preserve global information, maintain inter-block continuity, and establish building-level correspondence between exterior structures and interior spaces. We refer to this evolving cross-scale representation as the "living context".

During generation, the world context $\mathcal{C}_t$ provides task-specific conditions and evolves by incorporating validated results. For a generation step with task type $\tau_t$, the required conditions $\mathcal{W}_t$ are extracted from the current context through projection $\Pi_{\tau_t}$. The corresponding generator produces a candidate result $Y_t$ and newly derived evidence $\Delta_t$. Only validated results are integrated into the context:
\begin{equation}
\begin{aligned}
\mathcal{W}_t
&= \Pi_{\tau_t}(\mathcal{C}_t),\\
(Y_t,\Delta_t)
&= \mathcal{A}^{\tau_t}(\mathcal{W}_t).
\end{aligned}
\end{equation}
\begin{equation}
\begin{aligned}
\mathcal{C}_{t+1}
=
\begin{cases}
\mathcal{C}_t\oplus(Y_t,\Delta_t),
& \mathcal{Q}_{\tau_t}(\mathcal{C}_t,Y_t,\Delta_t)=\mathrm{pass},\\
\mathcal{C}_t,
& \text{otherwise}.
\end{cases}
\end{aligned}
\label{eq:checked-context-transition}
\end{equation}
Here, $\Pi_{\tau_t}$ selects the semantic, visual, spatial, and content information required by the current task, while $\oplus$ updates the corresponding context level with validated evidence. As generation proceeds, exterior synthesis queries city-level, block-level, and neighboring-region information, whereas indoor synthesis retrieves building-level semantics, appearance, assets, and geometric constraints. This continuous update mechanism ensures that only reliable information propagates through subsequent stages, maintaining consistency across the generated urban world.

\subsection{Exterior Realization and Building-Level Context Localization}

The city and block-level contexts are first realized as the urban exterior and then localized to individual buildings. Block generation establishes exterior information through context bridging, while building association further transfers this context to specific building instances and footprints, forming the building-level context $\mathcal{C}^{b}$ for indoor generation.

\subsubsection{Hierarchical Urban Planning and Style Specification}

Given the input description $P$, the City Planning Module defines the city theme, functional organization, and inter-block relationships over grid $G$, producing block-level plans $\{p_i\}$. It also derives shared exterior and interior style references to maintain a consistent visual identity across domains. The resulting planning and style information are incorporated into $\mathcal{C}^{\mathrm{world}}$.

Conditioned on $\mathcal{C}^{\mathrm{world}}$ and $p_i$, the Block Design Module generates block-specific spatial layouts that are stored in $\mathcal{C}^{\mathrm{block}}_i$ for external generation.

\subsubsection{Context-Aware Autoregressive Exterior Generation}

The exterior is generated autoregressively over blocks. For block $i$, the Exterior Generator is conditioned on world-level context, block-level design, and previously generated neighboring blocks:
\begin{equation}
I_i
=
\mathcal{A}^{\mathrm{out}}
\left(
\Pi_{\mathrm{out}}
\left(
\mathcal{C}^{\mathrm{world}},
\mathcal{C}^{\mathrm{block}}_i
\right),
\left\{I_j\right\}_{j\in\mathcal{N}^{-}_i}
\right).
\label{eq:autoregressive-exterior}
\end{equation}
where $\mathcal{N}^{-}_i$ denotes previously generated neighboring blocks. This autoregressive conditioning preserves spatial and visual continuity across block boundaries. After generation, each block image $I_i$ is converted into a 3D model $X_i$, and the resulting exterior $\mathcal{S}^{\mathrm{out}}$ is assembled over grid $G$.

\subsubsection{Building Instance Association and Geometric Grounding}

To transfer exterior context to indoor generation, we associate a building region $M_b^{\mathrm{iso}}$ in $I_i$ with its corresponding 3D instance $x_b\in X_i$. The building identity and footprint are obtained as:
\begin{equation}
\begin{aligned}
b&=(i,\operatorname{id}(x_b)),\\
F_b&=\operatorname{Footprint}(x_b).
\end{aligned}
\end{equation}

The building-level context is then constructed as:
\begin{equation}
\mathcal{C}^{b}
=
\Pi_{\mathrm{sv}}
(
\mathcal{C}^{\mathrm{world}},
\mathcal{C}^{\mathrm{block}}_i
)
\oplus
\{
I_i,M_b^{\mathrm{iso}},
\operatorname{id}(x_b),F_b
\}.
\label{eq:building-grounding}
\end{equation}
The resulting context integrates inherited semantics, appearance, instance identity, and geometric constraints, providing building-specific conditions for subsequent indoor synthesis.

\subsection{Building-Level Context-Driven Indoor Generation}

The building-level context $\mathcal{C}^{b}$ provides localized conditions for indoor synthesis by integrating inherited world information with building-specific appearance, identity, and geometry. Conditioned on $\mathcal{C}^{b}$, indoor generation derives building-specific content, assets, and spatial layouts while maintaining alignment with the corresponding exterior structure.

\subsubsection{Building-Specific Content Planning and Asset Generation}

Given $\mathcal{C}^{b}$, the indoor generation module first constructs a structured interior program that specifies functional spaces, spatial relationships, and required assets. Exterior appearance cues and inherited style information guide the generation of building-specific assets, forming an asset library $\mathcal{L}_b$ for subsequent synthesis.

The exterior appearance provides design evidence rather than direct observations of the hidden interior. By conditioning content planning and asset generation on $\mathcal{C}^{b}$, the resulting assets reflect both the building's functional role and its visual characteristics within the shared urban world.

\subsubsection{Exterior-Footprint-Constrained Hierarchical Interior Synthesis}

The recovered footprint $F_b$ defines the spatial domain of the target interior. Within this footprint, we perform hierarchical indoor synthesis from coarse spatial organization to fine-grained object placement. The process first determines room layouts, then progressively generates furniture, architectural elements, and smaller objects according to spatial dependencies. The process is formulated as:
\begin{equation}
\mathcal{S}^{\mathrm{in}}_b
=
\mathcal{A}^{\mathrm{in}}
\left(
\Pi_{\mathrm{in}}(\mathcal{C}^{b})
\right),
\qquad
\operatorname{Footprint}
\left(
\mathcal{S}^{\mathrm{in}}_b
\right)
\subseteq F_b,
\label{eq:hierarchical-interior}
\end{equation}
where $\mathcal{A}^{\mathrm{in}}$ denotes the hierarchical indoor synthesis process and $\Pi_{\mathrm{in}}$ extracts the interior program, style conditions, building-specific asset library $\mathcal{L}_b$, and footprint constraints from $\mathcal{C}^{b}$. The footprint constraint restricts the generated interior within the exterior building boundary, while hierarchical spatial dependencies regulate object placement. This formulation anchors indoor synthesis to its corresponding exterior building while preserving the contextual information inherited from the urban world.

\begin{table*}[t]
\centering
\setlength{\tabcolsep}{1mm}
\caption{Quantitative comparison of urban exterior generation. All methods receive identical urban descriptions as inputs. GPT-5.5-based and human evaluations are reported separately. The best results are in \textbf{bold}.}
\vspace{-0.25cm}
\label{tab:exterior-comparison}
\begin{tabular}{l*{8}{cc}}
\toprule
\multirow{3}{*}{Method} & \multicolumn{8}{c}{AQS} & \multicolumn{8}{c}{RDR} \\
\cmidrule(lr){2-9}\cmidrule(lr){10-17}
& \multicolumn{2}{c}{SVC$\uparrow$}
& \multicolumn{2}{c}{SRC$\uparrow$}
& \multicolumn{2}{c}{MTF$\uparrow$}
& \multicolumn{2}{c}{LA$\uparrow$}
& \multicolumn{2}{c}{SVC$\uparrow$}
& \multicolumn{2}{c}{SRC$\uparrow$}
& \multicolumn{2}{c}{MTF$\uparrow$}
& \multicolumn{2}{c}{LA$\uparrow$} \\
& GPT & Hum. & GPT & Hum. & GPT & Hum. & GPT & Hum.
& GPT & Hum. & GPT & Hum. & GPT & Hum. & GPT & Hum. \\
\midrule
CityCraft
& 7.11 & 7.17 & 7.22 & 6.00 & 4.89 & 5.00 & 5.89 & 5.00
& 23.65 & 19.21 & 20.76 & 19.61 & 19.90 & 14.99 & 17.35 & 15.97 \\
SynCity
& 7.70 & 7.17 & 8.20 & 7.83 & 4.90 & 6.33 & 5.30 & 6.50
& 21.95 & 18.27 & 25.69 & 21.72 & 16.97 & 15.12 & 14.91 & 12.00 \\
MajutsuCity
& 8.00 & 7.83 & 8.60 & 8.00 & 7.00 & 7.50 & 7.40 & 7.83
& 22.62 & 20.99 & 24.99 & 19.91 & 23.89 & 20.52 & 25.84 & 22.70 \\
\midrule
Ours
& \textbf{8.75} & \textbf{8.67}
& \textbf{9.00} & \textbf{8.50}
& \textbf{7.63} & \textbf{8.00}
& \textbf{8.00} & \textbf{8.67}
& \textbf{27.89} & \textbf{23.83}
& \textbf{27.64} & \textbf{24.65}
& \textbf{29.43} & \textbf{24.11}
& \textbf{29.57} & \textbf{25.89} \\
\bottomrule
\end{tabular}
\vspace{-0.1cm}
\end{table*}

\begin{table*}[t]
\centering
\setlength{\tabcolsep}{1mm}
\caption{Quantitative evaluation of building-level indoor-outdoor correspondence using GPT-5.5 and human evaluation. The best results are in \textbf{bold}. Dashes denote metrics unavailable for TRELLIS because its independently generated interior is not grounded in the exterior building footprint.}
\vspace{-0.25cm}
\label{tab:indoor-main}
\begin{tabular}{l*{6}{cc}c}
\toprule
\multirow{3}{*}{Method}
& \multicolumn{8}{c}{AQS}
& \multicolumn{4}{c}{RDR}
& \multirow{3}{*}{Shape IoU$\uparrow$} \\
\cmidrule(lr){2-9}\cmidrule(lr){10-13}
& \multicolumn{2}{c}{Functional$\uparrow$}
& \multicolumn{2}{c}{Visual$\uparrow$}
& \multicolumn{2}{c}{Spatial$\uparrow$}
& \multicolumn{2}{c}{Average$\uparrow$}
& \multicolumn{2}{c}{Functional$\uparrow$}
& \multicolumn{2}{c}{Visual$\uparrow$}
& \\
& GPT & Hum. & GPT & Hum. & GPT & Hum. & GPT & Hum. & GPT & Hum. & GPT & Hum. & \\
\midrule
TRELLIS & 6.23 & 6.30 & 5.88 & 4.85 & - & - & - & - & 19.17 & 20.03 & 1.83 & 10.90 & - \\
\midrule
Ours & \textbf{7.47} & \textbf{7.45} & \textbf{8.21} & \textbf{7.60} & \textbf{8.77} & \textbf{8.65} & \textbf{8.15} & \textbf{7.90} & \textbf{22.35} & \textbf{20.36} & \textbf{24.29} & \textbf{23.51} & \textbf{0.997} \\
\bottomrule
\end{tabular}
\vspace{-0.4cm}
\end{table*}

%% file: sec/experiment.tex
\section{Experiments}
\label{sec:experiments}

\subsection{Experimental Setup}

We evaluate our method on a diverse collection of generated cities covering different urban functions, architectural styles, and environmental themes. For urban exterior generation, we compare with CityCraft~\cite{citycraft}, SynCity~\cite{syncity}, and MajutsuCity~\cite{majutsucity}, using identical urban descriptions as inputs for all methods. Our framework supports arbitrary $R\times C$ block grids. For fair comparison, all experiments use $3\times3$ grids to maintain comparable city scales and inter-block relationships.

We use TRELLIS~\cite{trellis} as an independent-generation baseline. Given matched functional and stylistic descriptions, TRELLIS generates an exterior building and an interior scene independently, allowing us to examine whether shared textual conditions alone are sufficient to establish indoor-outdoor coherence.

For indoor evaluation, we randomly select valid buildings after instance association and generate their corresponding interior scenes. The system uses GPT-5.4 for language-based reasoning, GPT-Image-2 for block image generation and editing, and the Meshy API for image-to-3D conversion.

\subsection{Evaluation Metrics.}

We evaluate the proposed framework from three perspectives: urban exterior quality, indoor-outdoor coherence, and cross-block continuity.

For urban exterior generation, we adopt Absolute Quantitative Scoring (AQS) and Relative Dimension Ranking (RDR)~\cite{majutsucity}, following previous city generation evaluation protocols. Both protocols evaluate four dimensions: Structural and View Consistency (SVC), Scene Richness and Complexity (SRC), Material and Texture Fidelity (MTF), and Lighting and Atmosphere (LA). AQS assigns absolute scores ranging from 1 to 10, while RDR measures the relative preference between different methods through pairwise comparisons. We perform evaluations under identical criteria using both GPT-5.5-based assessments~\cite{gpteval3d,gen3deval} and human evaluations with 20 experts.

We measure indoor-outdoor coherence along three dimensions: functional, visual, and spatial consistency. Functional consistency evaluates whether the generated interior matches the semantic role of the target building. Visual consistency measures whether the interior preserves the building-specific appearance and the surrounding urban visual identity. Spatial consistency evaluates whether the interior layout conforms to the actual building footprint. In addition, we introduce Shape IoU as an evaluator-independent geometric metric to quantify the correspondence between the generated indoor envelope and the exterior building footprint.

Finally, we evaluate cross-block visual continuity to examine whether autoregressive neighborhood conditioning improves the coherence of generated cities. Specifically, we assess the continuity of road and ground surfaces, building appearance, and overall rendering style across adjacent blocks.

\begin{figure*}[th]
    \centering
    \includegraphics[width=1\linewidth]{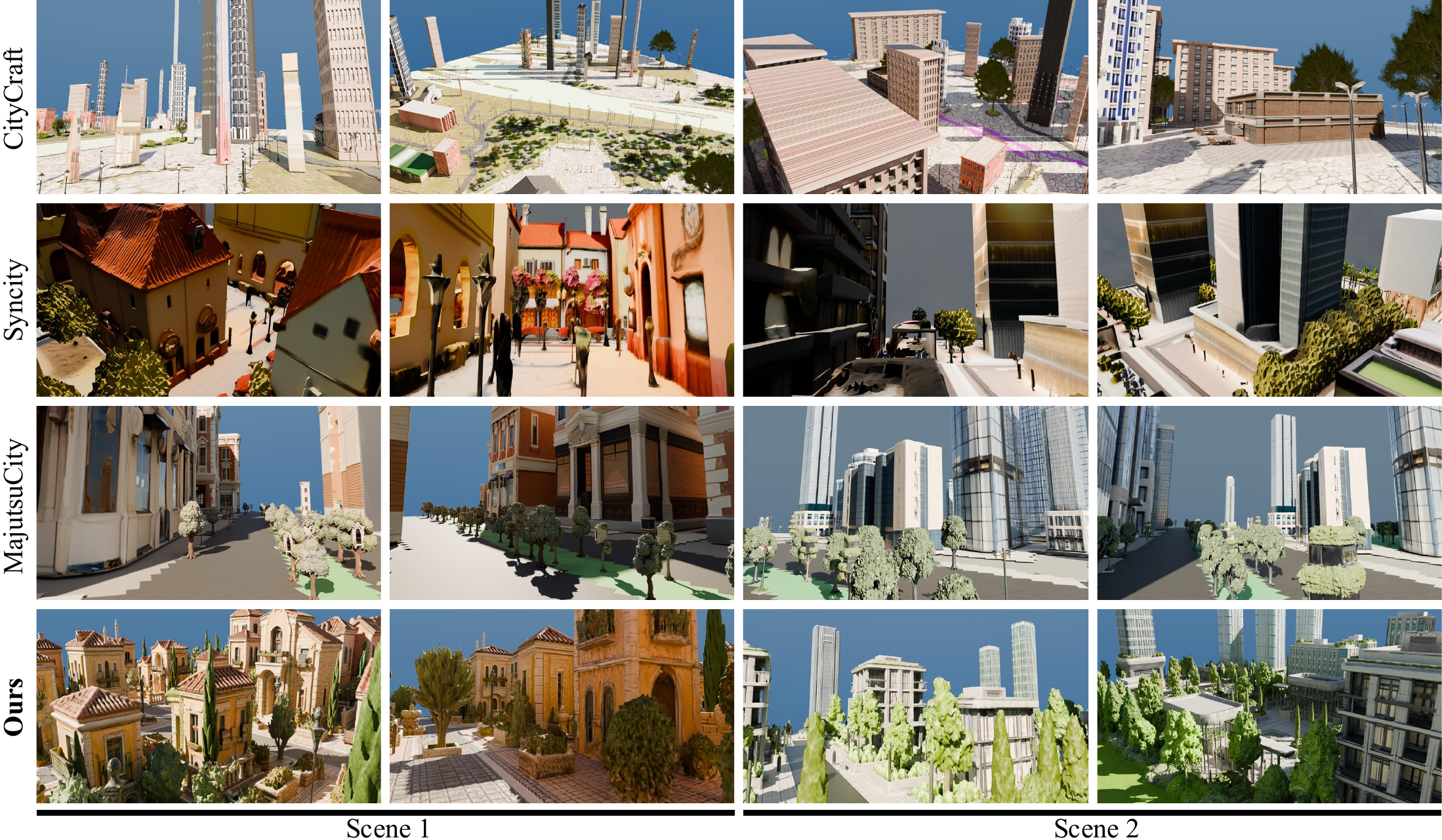}
    \caption{Qualitative comparison under identical urban descriptions, with two views per scene. HoloWorld generates richer architectural and landscape details while preserving more coherent spatial organization and visual style across the city.}
    \label{fig:exterior-comparison}
    \vspace{-0.35cm}
\end{figure*}

\subsection{Quantitative Comparison}

\subsubsection{Urban Exterior Generation.}

As shown in Table~\ref{tab:exterior-comparison}, our method achieves the best performance across all four evaluation dimensions under both AQS and RDR protocols. Under GPT-5.5-based evaluation, it improves the AQS scores over the strongest baseline MajutsuCity by relative margins of $9.38\%$, $4.65\%$, $9.00\%$, and $8.11\%$ on SVC, SRC, MTF, and LA, respectively. It also achieves the highest RDR scores in all four dimensions, indicating a stronger preference over competing approaches in pairwise comparisons. Human evaluation shows consistent results, with our method obtaining the highest AQS and RDR scores across all evaluated dimensions. These results demonstrate that our framework produces urban exteriors with improved structural coherence, richer scene composition, higher-fidelity appearance, and more consistent environmental styles.

\subsubsection{Building-Level Indoor-Outdoor Correspondence.}

As shown in Table~\ref{tab:indoor-main}, HoloWorld outperforms the paired-text independent TRELLIS baseline on all applicable functional and visual metrics under both GPT-5.5 and human evaluation. The largest gains appear in visual coherence: Visual AQS increases from $5.88$ to $8.21$ and RDR from $1.83$ to $24.29$ under GPT-5.5, with human judgments showing the same trend. HoloWorld additionally obtains Spatial AQS scores of $8.77$ and $8.65$ and a Shape IoU of $0.997$, reflecting explicit geometric grounding absent from the independent baseline. These results show that paired-text independent generation does not provide the functional, visual, and spatial correspondence required for a unified world.

\subsection{Qualitative Results}

\subsubsection{Urban Exterior Comparison.}

\begin{figure}[ht]
    \centering
    \includegraphics[width=1\linewidth]{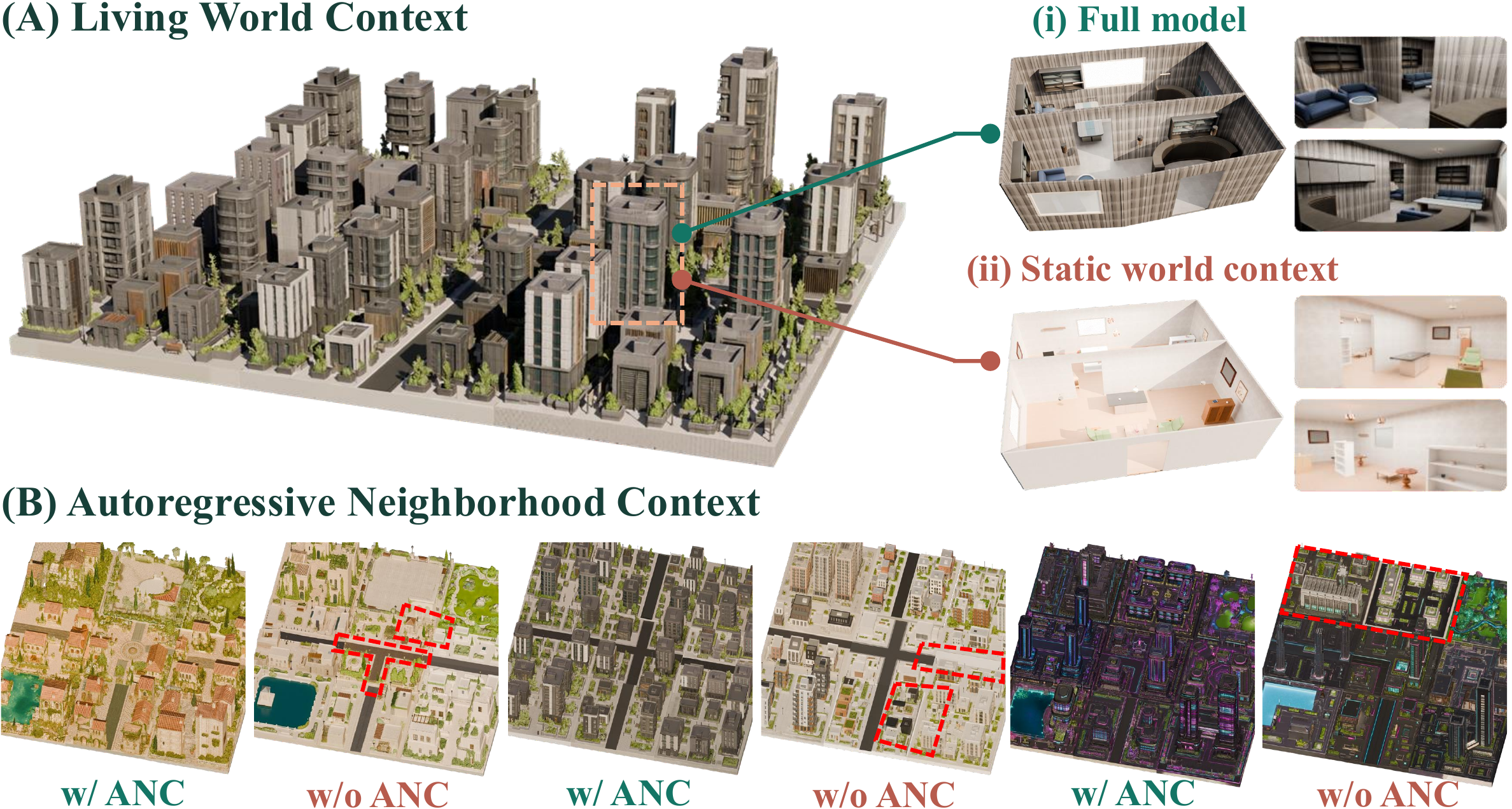}
    \caption{Qualitative results and ablations of HoloWorld. (A) The full model generates an interior that remains functionally, visually, and spatially consistent with its corresponding exterior building, whereas a static world context weakens this correspondence. (B) Autoregressive neighborhood conditioning preserves cross-block spatial and visual continuity; removing it introduces visible boundary discontinuities highlighted by the red boxes.}
    \label{fig:unified-generation}
    \vspace{-0.65cm}
\end{figure}

Figure~\ref{fig:exterior-comparison} presents representative urban scenes generated by all methods using identical urban descriptions. CityCraft and MajutsuCity produce well-structured urban layouts with strong geometric organization, but their results exhibit relatively less diversity in public spaces and fine-grained landscape elements. In contrast, our method jointly synthesizes urban infrastructures and elements (e.g., buildings, roads, parks, plazas, and even layered vegetation), producing richer urban environments with more distinctive regional characteristics. 

\begin{table*}[ht]
\centering
\small
\setlength{\tabcolsep}{1mm}
\caption{Effect of dynamic world-context localization on building-level indoor-outdoor correspondence under GPT-5.5 and human evaluation. The best results are in \textbf{bold}.}
\label{tab:context-update-ablation}
\vspace{-0.25cm}
\begin{tabular}{l*{7}{cc}c}
\toprule
\multirow{3}{*}{Configuration} & \multicolumn{8}{c}{AQS} & \multicolumn{6}{c}{RDR} & \multirow{3}{*}{Shape IoU$\uparrow$} \\
\cmidrule(lr){2-9}\cmidrule(lr){10-15}
& \multicolumn{2}{c}{Functional$\uparrow$}
& \multicolumn{2}{c}{Visual$\uparrow$}
& \multicolumn{2}{c}{Spatial$\uparrow$}
& \multicolumn{2}{c}{Average$\uparrow$}
& \multicolumn{2}{c}{Functional$\uparrow$}
& \multicolumn{2}{c}{Visual$\uparrow$}
& \multicolumn{2}{c}{Spatial$\uparrow$}
& \\
& GPT & Hum. & GPT & Hum. & GPT & Hum. & GPT & Hum.
& GPT & Hum. & GPT & Hum. & GPT & Hum. & \\
\midrule
Static World Context
& 5.75 & 5.88
& 4.58 & 6.00
& 3.25 & 4.75
& 4.53 & 5.54
& 16.37 & 11.75
& 1.40 & 0.33
& 1.40 & 11.09
& 0.670 \\
Full Model
& \textbf{7.67} & \textbf{7.88}
& \textbf{8.17} & \textbf{8.25}
& \textbf{7.75} & \textbf{8.50}
& \textbf{7.86} & \textbf{8.21}
& \textbf{21.82} & \textbf{18.86}
& \textbf{22.85} & \textbf{18.82}
& \textbf{22.85} & \textbf{17.50}
& \textbf{0.994} \\
\bottomrule
\end{tabular}
\vspace{-0.25cm}
\end{table*}

\begin{table}[ht]
\centering
\caption{Effect of autoregressive neighborhood conditioning (ANC) on cross-block visual and spatial continuity under GPT-5.5 and human evaluation. The best results are in \textbf{bold}.}
\label{tab:autoregressive-ablation}
\vspace{-0.25cm}
\begin{tabular}{lcccc}
\toprule
\multirow{2}{*}{Configuration}
& \multicolumn{2}{c}{Continuity AQS$\uparrow$}
& \multicolumn{2}{c}{RDR$\uparrow$} \\
\cmidrule(lr){2-3}\cmidrule(lr){4-5}
& GPT & Hum. & GPT & Hum. \\
\midrule
Full Model & \textbf{8.25} & \textbf{7.75} & \textbf{24.04} & \textbf{21.97} \\
w/o ANC & 7.25 & 5.38 & 16.66 & 19.07 \\
\bottomrule
\end{tabular}
\vspace{-0.35cm}
\end{table}

Compared with the tile-based generation strategy of SynCity, our method achieves more coherent urban organization and visual consistency across neighboring blocks while preserving detailed local structures in buildings, roads, and landscape components. These results demonstrate that our framework can generate not only visually rich urban environments but also globally organized and spatially coherent city-scale scenes.

\subsubsection{Unified Indoor-Outdoor Generation.}

Figure~\ref{fig:unified-generation}(A) pre-sents representative results of our unified generation framework, where scenes are progressively instantiated from city layouts and block exteriors to individual building interiors. Each generated interior is explicitly grounded to a specific exterior building through building-level correspondence: its spatial organization follows the building footprint, its functional layout matches the semantic role of the building, and its assets and materials inherit the visual identity of the building and the surrounding urban context. Different buildings exhibit distinct interior organizations and content compositions that are adapted to their functions, appearances, and spatial constraints, rather than relying on a shared indoor template. These results demonstrate that our method generates interiors as coherent continuations of the same outdoor urban world instead of independent indoor scenes.

\subsection{Ablation Studies}

We assess dynamic world-context updates and autoregressive neighborhood conditioning through \textbf{Static World Context} and \textbf{w/o ANC}, targeting building-level indoor-outdoor correspondence and cross-block continuity, respectively.

\textbf{Static World Context} and the full model target the same exterior building and generate the same number of rooms to control scene complexity. Unlike the full model, the static configuration receives only the initial city description without any subsequent context updates and building-level localization. Under this setting, the model predicts room functions, contents, and assets from the initial description while using a default rectangular boundary for spatial layout. 

As shown in Table~\ref{tab:context-update-ablation}, freezing dynamic world-context updates substantially degrades building-level indoor-outdoor correspondence. Under GPT-5.5 evaluation, freezing the context reduces the functional, visual, and spatial AQS scores from $7.67$, $8.17$, and $7.75$ to $5.75$, $4.58$, and $3.25$, respectively. Meanwhile, Shape IoU decreases from $0.994$ to $0.670$, indicating that the generated interiors no longer accurately follow the corresponding building geometry. Human evaluation shows the same trend, with the average AQS decreasing from $8.21$ to $5.54$ without context updates. These results demonstrate that the initial city description alone cannot provide sufficient building-specific semantic, visual, and geometric constraints, whereas the continuously updated and localized world context enables interiors to remain traceable to their corresponding exterior buildings.

\textbf{Without Autoregressive Neighborhood Context}, each block retains the same city-level description, block-level design, and style conditions as the full model but does not access previously generated neighboring blocks. This setting isolates the effect of autoregressive neighborhood conditioning on cross-block visual continuity. Figure~\ref{fig:unified-generation}(B) highlights the resulting boundary discontinuities, while Table~\ref{tab:autoregressive-ablation} shows that removing autoregressive neighborhood conditioning reduces cross-block continuity under both GPT-5.5-based and human evaluation. Under GPT-5.5 evaluation, the continuity AQS decreases from $8.25$ to $7.25$, while the RDR score drops from $24.04$ to $16.66$. Human evaluation exhibits the same trend, with AQS decreasing from $7.75$ to $5.38$ and RDR decreasing from $21.97$ to $19.07$. These results demonstrate that previously generated neighboring blocks provide valuable contextual references, enabling subsequent blocks to preserve local spatial relations and visual characteristics while maintaining global urban continuity.

%% file: sec/conclusion.tex
\section{Conclusion}
\label{sec:conclusion}
We present HoloWorld, a unified framework for indoor-outdoor urban scene generation that treats exteriors and interiors as corresponding realizations of the same 3D world. Built upon a continuously updated cross-scale world context, HoloWorld transfers and shares semantic, visual, and spatial information from city-level planning to individual buildings, enabling each generated interior to maintain explicit correspondence with its corresponding exterior. Meanwhile, context-aware autoregressive generation preserves spatial and visual continuity across neighboring urban blocks.

Experiments demonstrate that HoloWorld achieves superior urban exterior generation quality compared to existing city generation methods and maintains a strong building-level correspondence between generated interiors and their associated buildings. Ablation studies further verify the essential role of the living context and the importance of autoregressive neighborhood conditioning. These results show that HoloWorld bridges previously separated indoor and outdoor generation processes, enabling them to form coherent components of a unified 3D urban world.

Our current framework focuses on single-floor interiors and does not explicitly model vertical building structures. Future work will extend HoloWorld toward multi-floor architectural reasoning, richer structural constraints, and interactive physical validation, enabling more comprehensive urban-scale simulations and embodied-agent applications.